\documentclass[english,runningheads]{llncs}
\usepackage[utf8]{luainputenc}
\usepackage{babel}
\usepackage{graphicx}
\usepackage{tabularx}
\usepackage{multirow}
\usepackage{graphics}
\usepackage{amsmath}
\usepackage{hyperref}
\newcommand{\mytilde}{\raise.17ex\hbox{$\scriptstyle\mathtt{\sim}$}}

\begin{document}

\title{Single- and Multi-Task Architectures for Surgical Workflow Challenge
at M2CAI 2016}

\author{Andru P. Twinanda \inst{1} \and Didier Mutter\inst{2} \and Jacques Marescaux \inst{2} \and Michel De Mathelin \inst{1} \and Nicolas Padoy \inst{1}}

\authorrunning{ }
\titlerunning{ }

\institute{ICube, University of Strasbourg, CNRS, IHU Strasbourg, France \\ \and IRCAD, IHU Strasbourg, University Hospital of Strasbourg, France \\}

\maketitle

\section{Introduction}

\begin{figure}
\begin{centering}
\includegraphics[width=12cm]{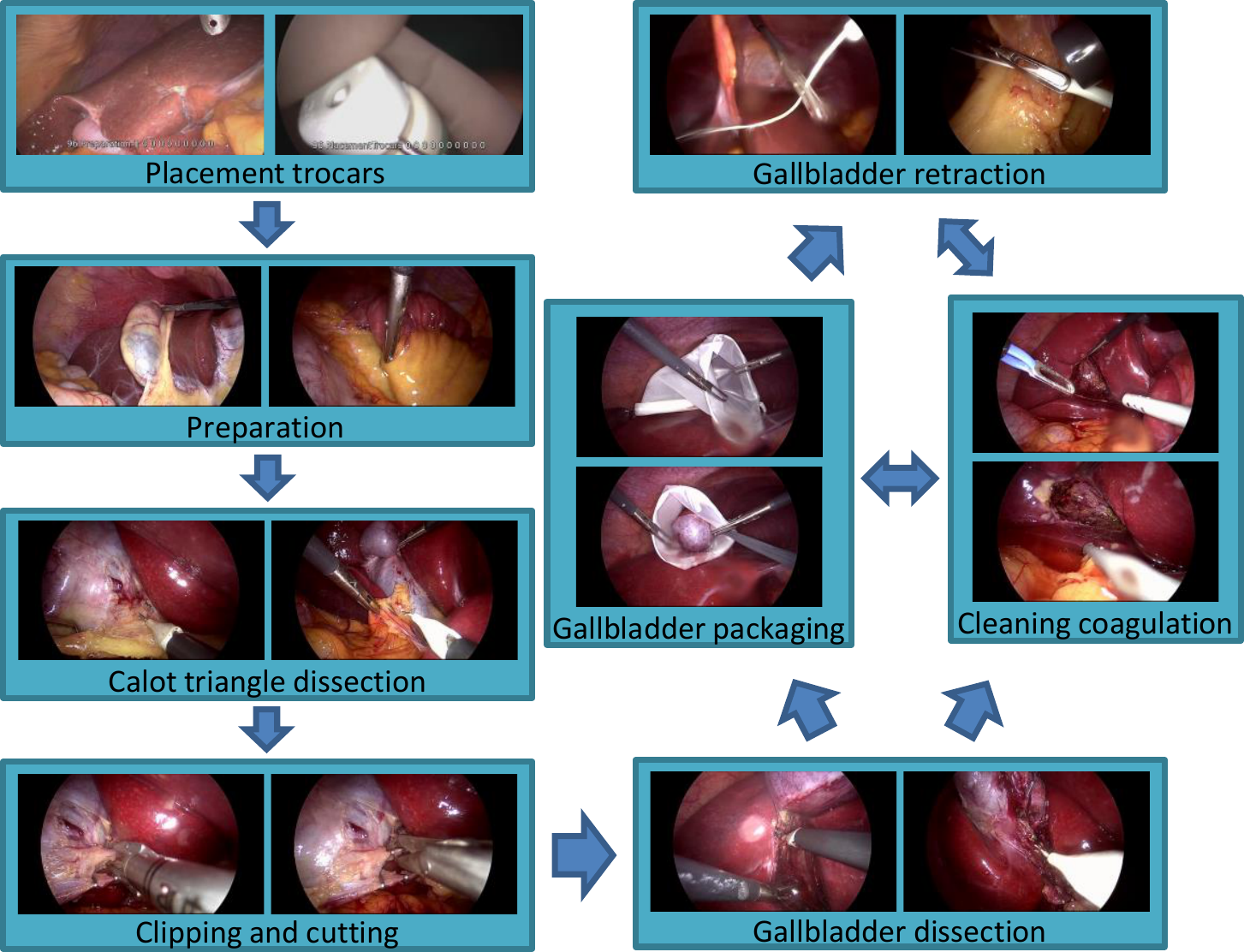}
\par\end{centering}
\caption{Defined phases and their transitions found in the m2cai2016-workflow dataset. \label{fig:defines-phases}}
\end{figure}

The surgical workflow challenge at M2CAI 2016 consists of identifying
8 surgical phases in cholecystectomy procedures. In Fig. \ref{fig:defines-phases},
we show the defined phases as well as the phase transitions observed
in the \textit{m2cai2016-workflow} dataset \cite{ostler_healthcom2015,twinanda_tmi2016}\footnote{The dataset is available at the official web page of M2CAI 2016: \url{http://camma.u-strasbg.fr/m2cai2016/} }. The training dataset,
released on May 23, 2016, consists of 27 cholecystectomy videos annotated
with the phases at 25 fps; while the testing dataset, released on
September 9, 2016, consists of 14 videos.

\begin{figure}
\begin{centering}
\begin{tabular}{cc}

\multicolumn{2}{c}{\includegraphics[width=12cm]{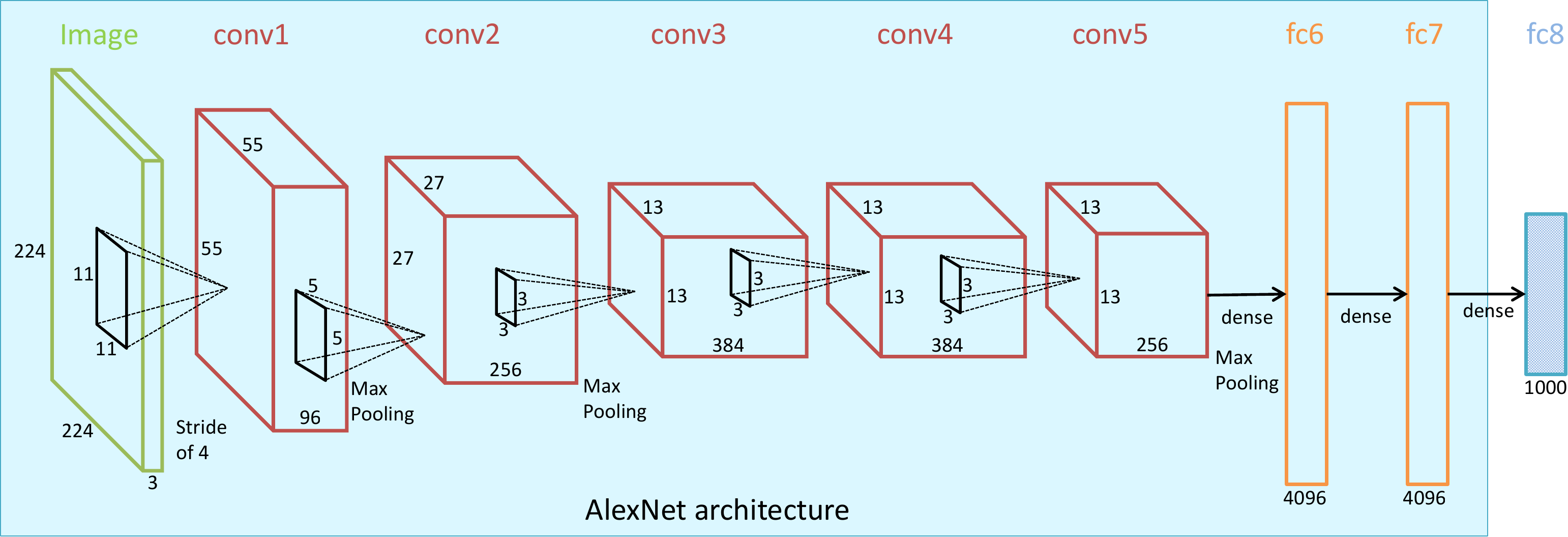}}\\
\multicolumn{2}{c}{(a)}\\

\includegraphics[height=2cm]{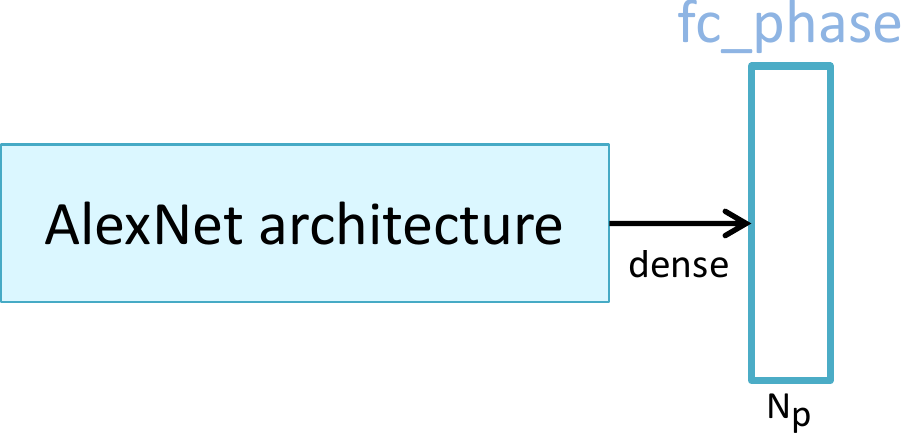} & \includegraphics[height=2cm]{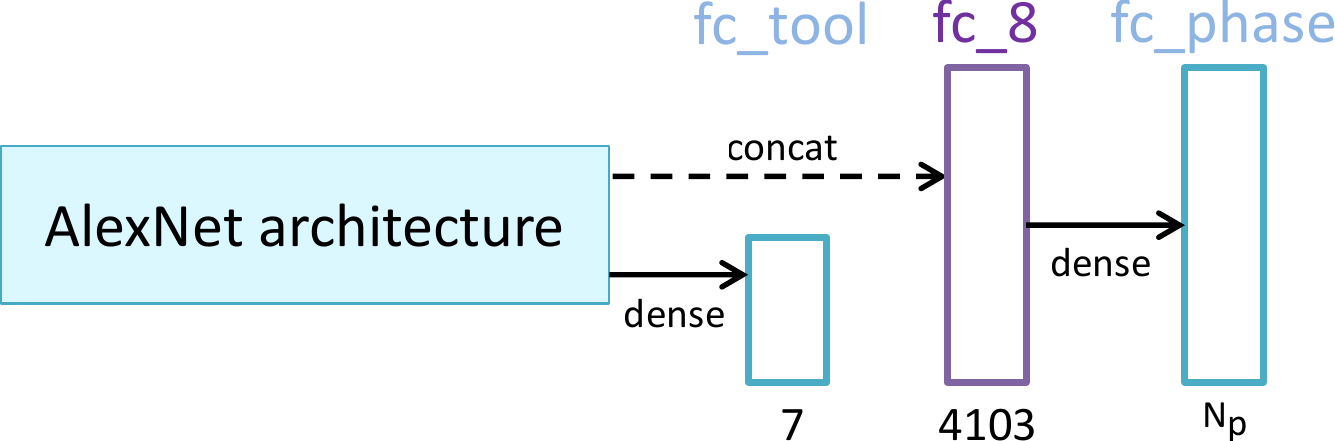} \\
(b) & (c) 
\end{tabular}
\par\end{centering}
\caption{The architectures of: (a) AlexNet, (b) PhaseNet, and (c) EndoNet. $N_p$ is the number of phases in the dataset used to finetune the networks. \label{fig:network-architectures}}
\end{figure}

Here, we propose to use deep architectures to perform the phase recognition
task. This work is based on our previous work \cite{twinanda_tmi2016}
where we presented several network architectures to perform multiple recognition tasks on laparoscopic videos. The tasks are surgical phase recognition and tool presence detection. Ultimately, we proposed an architecture which is designed to jointly perform both tasks. In this work, we are using both single-task and multi-task networks to learn the discriminative visual features from the dataset.

Naturally, surgical procedures are performed accordingly to a pre-defined surgical workflow. Thus, to properly perform surgical phase recognition, it is important to enforce the temporal constraints coming from the surgical workflow. On the other hand, the networks only accepts images in a frame-wise manner, thus there is not any temporal information incorporated in the results given by the networks. Therefore, an additional pipeline is required to enforce these temporal constraints. In \cite{twinanda_tmi2016}, we enforce the surgical workflow constraint by using an approach based on Hidden Markov model (HMM). However, HMMs work under the Markov assumption where the current state only depends on the previous state. In addition, the number states passed along a sequence is typically limited to the number of classes defined in the problem. These limitations are however not present in long-short term memory (LSTM) network. In this work, we are also going to perform the surgical phase recognition task using a LSTM network and compare the recognition results to the ones obtained by the HMM pipeline. 

\section{Methodology}

In previous work \cite{twinanda_tmi2016}, we proposed two convolutional
neural network (CNN) architectures to perform surgical phase recognition: PhaseNet and EndoNet, shown in
Fig. \ref{fig:network-architectures}. PhaseNet
is designed to solely perform the phase recognition task, while EndoNet
is designed to jointly perform the phase recognition and tool presence
detection tasks.  In \cite{twinanda_tmi2016}, it has been shown that the multi-task
network performs better than the single-task counterpart. However,
the multi-task network requires both phase and tool presence annotations
which are not available in the m2cai16-workflow dataset. In Section \ref{sect:feature-comparison}, we will explain how we conduct our experiments to cope with this limitation.

Note that the network is finetuned to perform the phase recognition
task using solely image features, thus there is no temporal constraint incorporated in the prediction process. In order to enforce the
temporal constraints, we propose to use two different approaches:
(1) HMM-based and (2) LSTM-based. The HMM-based approach is similar
to the one presented in \cite{twinanda_tmi2016}. First, we extract
image features (the output of the second last layer of each network, i.e., $\mathtt{fc7}$ in PhaseNet and $\mathtt{fc8}$ in EndoNet) from the
video frames. Then, they are passed to a multi-class linear SVM to
compute the values representing the confidences of an image belonging
to the phases. Ultimately, these confidences are then taken as input
to a hierarchical HMM (HHMM). Since the recognition is performed online,
we use the forward algorithm to compute the final predictions.

\begin{figure}[t!]
\begin{centering}
\includegraphics[width=10cm]{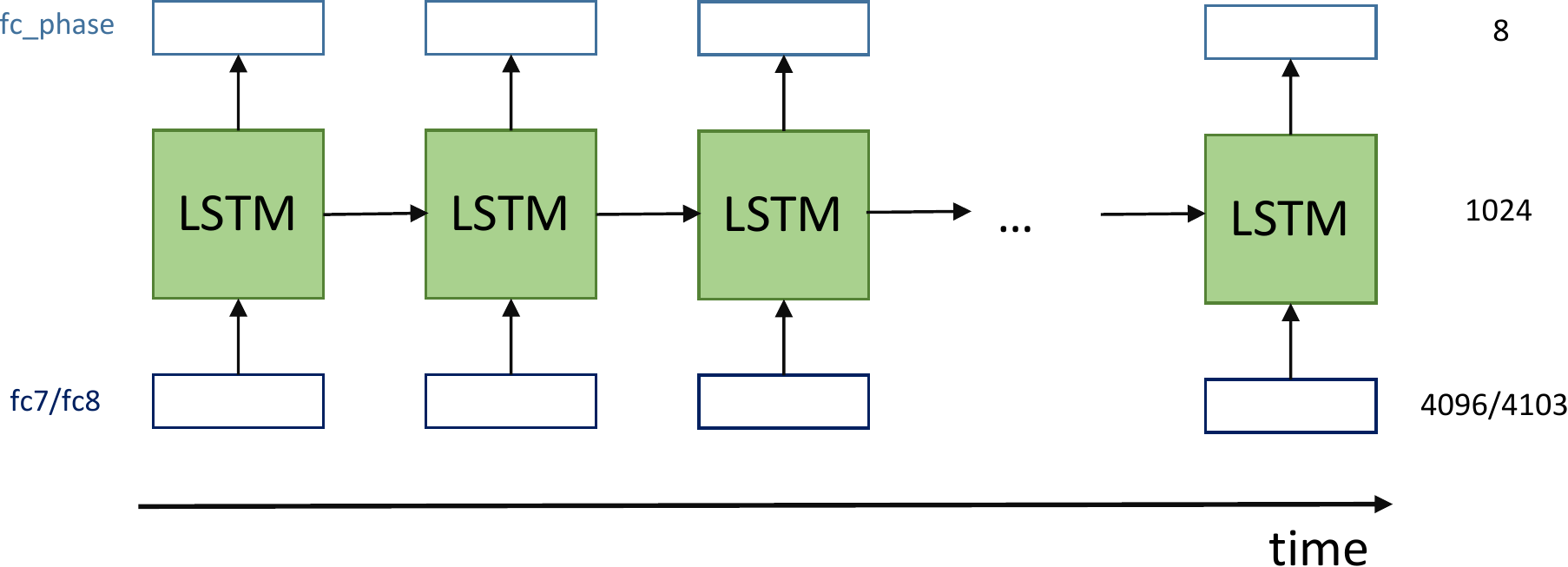}
\par\end{centering}
\caption{LSTM architecture for phase recognition. \label{fig:lstm-architecture}}
\end{figure}

The second approach uses long-short term memory (LSTM) network to
enforce the temporal constraint. We pass the image features to an
LSTM network with 1024 states. These states are then passed to a fully
connected layer with 8 nodes (equal to the number of phases in the m2cai16-workflow dataset). The output values of this fully connected
layer represent the confidences of the image belonging to the phases
and are used for final predictions. The LSTM network is shown in Figure \ref{fig:lstm-architecture}.

\section{Experimental Setup}

\subsection{Feature Comparisons \label{sect:feature-comparison}}

As previously stated, the EndoNet architecture is designed to perform jointly surgical phase recognition and tool presence detection while the m2cai16-workflow does not contain tool binary annotations. To cope with this limitation, we are using the Cholec80 dataset \cite{twinanda_tmi2016} which contains both phase and tool binary annotations. In addition to the additional annotations, the Cholec80 dataset contains more training videos than the m2cai16-workflow dataset (i.e., 40 vs. 27 training videos). However, the phase definition in Cholec80 is not the same as the one in m2cai16-workflow (7 vs. 8 phases).  Thus, the number of nodes in the $\mathtt{fc\_phase}$ has to be adjusted accordingly with respect to the datasets used to finetune the network. Here, we will finetune multiple networks with the PhaseNet and EndoNet architectures using m2cai16-workflow and Cholec80.

In summary, we are going to compare the performances of the following networks:
\begin{itemize}
\item PhaseNet-m2cai16. This network is trained using the PhaseNet architecture on the m2cai16-workflow dataset ($N_p=8$);
\item PhaseNet-Cholec80. This network is trained using the PhaseNet architecture on the Cholec80 dataset ($N_p=7$);
\item EndoNet-Cholec80. This network is trained using the EndoNet architecture on the Cholec80 dataset ($N_p=7$).
\end{itemize}

\subsection{PhaseNet and EndoNet Finetuning Parameters}

All networks are trained by fine-tuning the publicly available AlexNet network \cite{krizhevsky_nips2012}, which has been pre-trained on the ImageNet dataset \cite{imagenet}. The layers that are not defined in AlexNet (i.e., $\mathtt{fc}$\_$\mathtt{tool}$ and $\mathtt{fc}$\_$\mathtt{phase}$) are initialized randomly. 
The network is fine-tuned for 50K iterations with $N_{i}=50$ images in a batch.
The learning rate is initialized at $10^{-3}$ for all layers, except
for $\mathtt{fc}$\_$\mathtt{tool}$ and $\mathtt{fc}$\_$\mathtt{phase}$,
whose learning rate is set higher at $10^{-2}$ because of their random initialization. The learning rates for all layers decrease by a factor of
$10$ for every 20K iterations. The fine-tuning process is carried
out using the Caffe framework \cite{caffe}. 


\subsection{Phase Recognition Pipeline}

The phase recognition pipeline is trained to enforce the temporal constraints into the recognition process. Thus, it is important to note that the following approaches are solely trained using the m2cai16-workflow dataset.\\
\\
\textbf{HMM-based pipeline.} To carry out phase recognition, all image features (i.e., second last layer of respective network) are passed to a one-vs-all \textit{linear} SVM. For the HHMM, we set the number of top-level states to eight (equal to the number of phases in m2cai16-workflow), while the number of bottom-level states is data-driven (as in \cite{padoy_cvw2009}). To model the output of the SVM, we use a mixture of five Gaussians for every feature, except for the binary tool signal, where one Gaussian is used. The type of covariance is diagonal.\\
\\
\textbf{LSTM-based pipeline.} Due to memory constraints, it is still difficult to train the CNN and the LSTM networks in an end-to-end manner since each video typically lasts more than 30 minutes. In order to solve this problem, we train the CNN and LSTM networks separately. To do so, first we extract the image features using the finetuned networks (both PhaseNet and EndoNet) and train the LSTM pipeline using these extracted features. The LSTM network is trained over complete sequences using one video per batch. Each sequence comprises 3993 frames, which corresponds to the maximum video duration found in the dataset, i.e., 3993 seconds since we are working at 1 fps. For videos that are shorter than 3993 seconds, we pad the sequences with zeros. Since the LSTM is not finetuned on a pre-trained network, we set the learning rates to $10^{-2}$. The LSTM pipeline training process is carried out using the Caffe framework \cite{caffe} and it is performed for 30K iterations.

\subsection{Evaluation Metrics}

The surgical workflow challenge is evaluated using the Jaccard score, which is computed as follows:
\begin{equation}
J(GT,P) = \dfrac{GT \cap P}{GT \cup P},
\end{equation}
where $GT$ and $P$ are respectively the ground truth and prediction for each phase. In addition to that, we will also show the accuracy of the methods.

\section{Experimental Results}

\begin{table}[t!]
\begin{centering}
\begin{tabular}{|c|c|c|c|c|}
\hline
\multirow{2}{*}{Network} & \multicolumn{2}{c|}{HMM} & \multicolumn{2}{c|}{LSTM} \\
 \cline{2-5} &  Jaccard & Accuracy & Jaccard & Accuracy \\
\hline
\hline
PhaseNet-m2cai16 & 64.1$\pm$10.3 & 79.5$\pm$12.1 & 54.8$\pm$8.9\;\; & 72.5$\pm$10.6 \\
\hline
PhaseNet-Cholec80 & 62.4$\pm$10.4 & 71.1$\pm$20.3 & 64.4$\pm$10.0 & 80.7$\pm$12.9 \\
\hline
EndoNet-Cholec80 & 67.7$\pm$10.9 & 80.6$\pm$11.5 & 69.8$\pm$7.1\;\; & 80.1$\pm$17.6\\
\hline
\end{tabular}
\par\end{centering}
\caption{Phase recognition results. \label{tab:recognition-results}}
\end{table}

We show the phase recognition results in Table \ref{tab:recognition-results}. Using the HMM-based pipeline, despite the increase of training size, PhaseNet-Cholec80 does not necessarily perform better than PhaseNet-m2cai16. This might be due to the fact that PhaseNet-Cholec80 is trained on a dataset which contains a different phase definition to the one in m2cai16-workflow. Thus, the extracted features are not finetuned to perform the objective of this challenge. However, it is interesting to see that this is not observed in the results of the multi-task network (EndoNet-Cholec80). Even though it has not been trained on m2cai16-workflow, EndoNet-Cholec80 outperforms the PhaseNet-m2cai16. This is in line with the conclusion from \cite{twinanda_tmi2016} that finetuning the network in a multi-task manner will result in a better network for the phase recognition task. 

We also show the results of the LSTM-based pipeline in Table \ref{tab:recognition-results}. One can observe that there is an improvement of performance when PhaseNet-Cholec80 and EndoNet-Cholec80 are used. On the other hand, there is a drop of performance when the recognition is performed using the PhaseNet-m2cai16 features. This drop of performance might occur due to the fact that we set the LSTM hyperparameters equal to what we had found in our previous experiments with the Cholec80 dataset, yet these hyperparameters might result in bad performance on the m2cai16-workflow dataset. Due to time constraint, we are unable to thoroughly perform the hyperparameter search for this challenge. We believe that by properly tuning the hyperparameters, the LSTM results could be further improved. 

\section{Conclusions}

We have presented several approaches to perform surgical phase recognition for the surgical workflow challenge at M2CAI 2016. We proposed to use two types of CNN architectures to address the task: PhaseNet and EndoNet. The former performs the phase recognition task in a single-task manner, while the latter performs the task jointly with the tool presence detection task. The results show that the features extracted from a multi-task network perform better than the ones from a single-task one. From the results, we can also see that the LSTM-based approach was able to outperform the HMM-based approach and also to properly enforce the temporal constraints into the recognition process.

Here, the CNN and the LSTM trainings are performed separately. In order to establish an end-to-end architecture, it would be interesting to train them jointly. However, this is not a trivial task since it requires a lot of memory to train the network in an end-to-end manner.

\bibliographystyle{plain}
\bibliography{bibliography}

\end{document}